# Machine learning-based condition monitoring of powertrains in modern electric drives

Dinan Li, Panagiotis Kakosimos, and Luca Peretti

*Abstract*— The recent technological advances in digitalization have revolutionized the industrial sector. Leveraging data analytics has now enabled the collection of deep insights into the performance and, as a result, the optimization of assets. Industrial drives, for example, already accumulate all the necessary information to control electric machines. These signals include but are not limited to currents, frequency, and temperature. Integrating machine learning (ML) models responsible for predicting the evolution of those directly collected or implicitly derived parameters enhances the smartness of industrial systems even further. In this article, data already residing in most modern electric drives has been used to develop a data-driven thermal model of a power module. A test bench has been designed and used specifically for training and validating the thermal digital twin undergoing various static and dynamic operating profiles. Different approaches, from traditional linear models to deep neural networks, have been implemented to emanate the best ML model for estimating the case temperature of a power module. Several evaluation metrics were then used to assess the investigated methods' performance and implementation in industrial embedded systems.

## I. INTRODUCTION

Today, an industrial drive is much more than a device that produces the necessary output signals to adjust the motor speed and torque; it constitutes the core of the powertrain. An electric powertrain may comprise several components, including the drive, motor, storage means, transmission, and conveyor belt (Figure 1). Signals for controlling an asset usually contain insightful information about not only the specific asset but also the rest powertrain components and even the application. Thus, a drive gathers plenty of information about a system's operation, making it feasible to identify and eliminate possible limiting factors and optimize the whole powertrain's performance. In the past, a drive's primary task was to perform control actions; however, in modern industrial systems, drives play a more vital role by being equipped with several sensors and powerful microprocessors. The physical assets may be surrounded by several IoT peripherals that link to cloud services or support the execution of artificial intelligence (AI) algorithms on the edge [1]. The possibility of utilizing the vast amount of data residing in an electric drive and processing them locally or remotely opens new prospects in industrial digitalization [2].

Monitoring the thermal behavior of an asset by estimating its temperature under various operating scenarios has been one of the first applications of ML/AI. Today, ML models offer unmatched performance over classic thermal modeling approaches without requiring sophisticated hardware architectures and enormous resources [3]. Temperature signals, previously utilized as threshold limits under heavy-duty conditions in industrial systems, can now be used to monitor and optimize their performance under any operating condition. Thermal design and temperature ratings of powertrain components can benefit by retrofitting the thermal model output to design and dimensioning tools [4]. Under faulty conditions, an adaptive model may inherently attempt to forecast the device temperature; however, proper adaptations may support the detection of deviations from the expected trajectory and signal an alert [5].

It is thus evident that estimating or forecasting the temperature of an asset is crucial for energy saving, performance optimization, condition monitoring, and lifetime extension, among others [6]–[8]. Establishing the relationship of power losses with temperature variation has been studied and led to the development of advanced analytical or computational thermal models [9]. However, precise knowledge of the power losses for estimating changes in thermal behavior is necessary. Identifying the losses during the system operation while several unexpected factors influence its behavior is a challenging and sometimes unsolved task [10]. Therefore, static thermal models under known operating conditions are usually employed in the end. It is thus essential to develop a model to track real-time temperature changes and define the relationship between measured signals and temperatures while inputting all the necessary information that impacts thermal behavior [11].

Dinan Li and Panagiotis Kakosimos are with the Corporate Research Center, ABB AB, 72358 Västerås, Sweden (e-mail: Panagiotis.Kakosimos@se.abb.com).

Luca Peretti is with the Division of Electric Power and Energy Systems, KTH Royal Institute of Technology, 10044 Stockholm, Sweden (e-mail: lucap@kth.se)

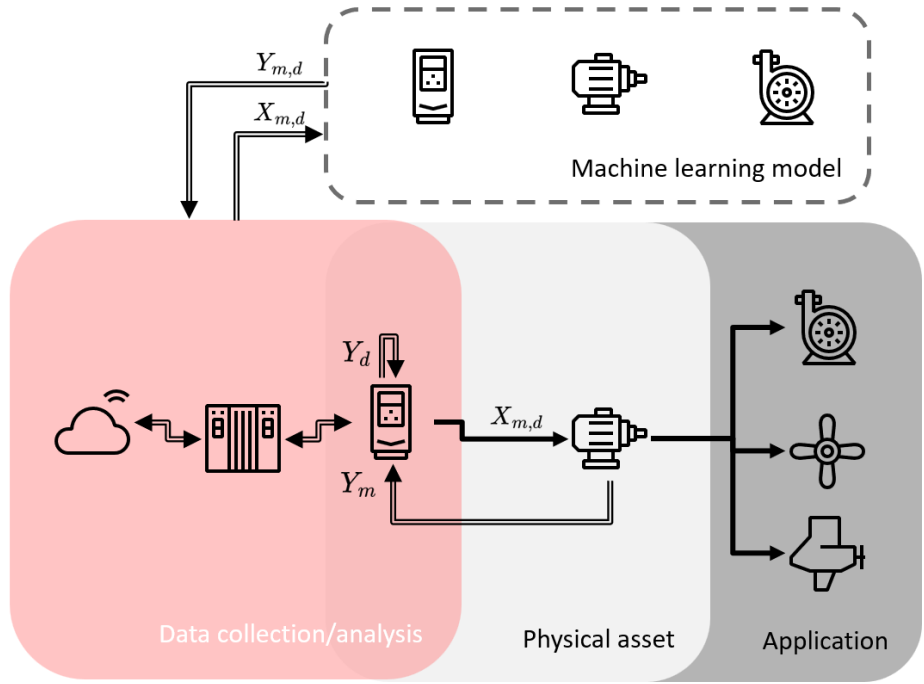

Figure 1. Block diagram of a modern electric powertrain.

An ML model can be used to establish this connection by understanding how already measured operating parameters are correlated with temperature changes. Adding more layers to a neural network structure may further assist in considering even more abstract phenomena without explicitly defining their interactions with the temperature. This article explores the use of existing measured signals in today's electric drives to estimate the case temperature of power modules operating under static and dynamic conditions. The accurate estimation of power losses is not required because the ML model can account for their impact on the temperature variation directly from the measured signals. The case temperature of a power module has been used as an illustrative example; nevertheless, similar results and conclusions can be drawn for estimating the thermal behavior of motors or any other powertrain components.

## II. SPECIFICATIONS OF THE MACHINE LEARNING MODEL

The development of any ML model requires the utilization of a pipeline that assembles several steps and cross-validates them together by setting different parameters (Figure 2). The first step of such a pipeline is the data collection and preparation for training the ML model. The main goal is to develop a data-driven thermal model of a power module, so the drive must undergo several speed and load conditions. For this purpose, seventeen profiles of about eight hours each with both fast and slow dynamics have been created for the investigated electric powertrain (Table 1). During the execution of each profile, several internal drive parameters have been captured with a sampling frequency of about 1 Hz. At this point, the cross-validation of all the captured parameters dictates the ones that must be added to the model because they influence the power module temperature. The selected parameters (Table 2) are directly related to the temperature changes, whereas the ambient temperature is needed to provide the environmental conditions. In addition to those parameters, feature expansion has also been conducted by adding eight exponentially weighted moving averages for considering the past thermal behavior.

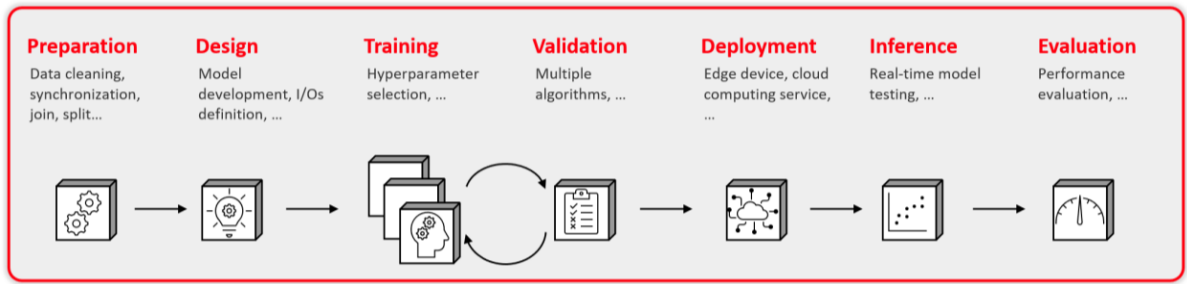

Figure 2. Process of model development.

In the next step, various ML models of different complexity have been tested, and their performances in estimating the case temperature have been evaluated by splitting the datasets to test, train, and validation. In machine learning, a loss or cost function is a function that is used to represent how well a model fits the data. The loss function is computed given the model's estimation and the measured real value. There are many candidates for loss functions, and they can be classified based on the nature of the targets. Since data collected from the field is necessary for training the ML model here, only supervised ML algorithms taking continuous target variables have been investigated. Several models have been benchmarked before selecting the most appropriate ones (Figure 3). These ML models are among the most prevalent for similar applications by being superior in terms of complexity and hardware resources:

Table 1. Specifications of the investigated setup

| Motors | Test motor |
|---|---|
| Type | Induction machine |
| Power rating | 15 kW |
| Voltage rating | 400 V |
| Current rating | 30.6 A |
| Torque rating | 97 N m |
| Pole pair number | 2 |
| Nominal speed | 1478 rpm |
| Cooling means | Attached fan blades |
| **Drives** | Test drive |
| Type | ACS880-01-038A |
| Topology | 2-level, IGBT |
| Power rating | 26 kVA |
| Current rating | 38 A |
| Cooling means | Forced air |

- **Linear regression (Elastic Net)** is the simplest form to be considered. This model assumes the existence of a linear relationship between the scalar response and one or more explanatory variables. Regularization has been used to penalize the terms of the loss function, thus preventing overfitting and improving the generalizability of the model.
- **A multilayer perceptron (MLP)**, or fully connected neural network, is the simplest feed-forward network. It often includes three layers: an input layer, a hidden layer, and an output layer, and each layer consists of many neurons. All the neurons between two consecutive layers are interconnected to make a fully connected neural network. The value of later neurons is a linear combination of the previously connected neurons by applying an activation function.
- **Convolutional Neural Networks (CNNs)** are famous for their application in image processing because of their capability to explore spatial data; however, the one-dimensional variant constitutes a good fit to explore temporal data. Unlike a fully connected network, data must come sequentially for a temporal CNN to work. Each neuron has information from not only the current time step but also from previous inputs. If more convolutional layers are stacked, neurons at the last layers can quickly get information from inputs that originated a long time ago.

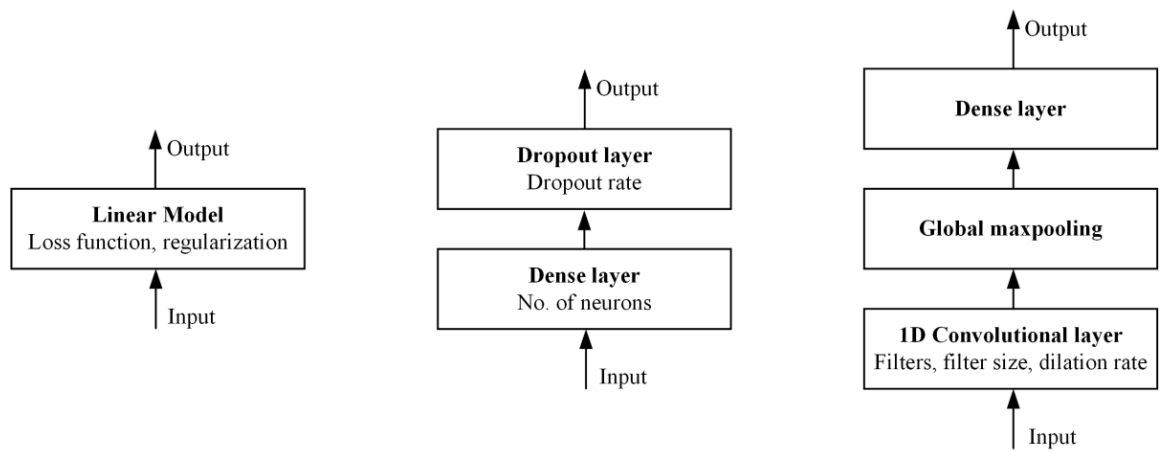

Figure 3. Block diagrams of Elastic Net, MLP, and CNNs, from right to left.

A critical step of the ML model development is tuning the hyperparameters. Some parts of the process can be fully automated using Bayesian optimization; however, other parts require careful consideration of the model behavior, thus consuming significant time and computational resources. Model validation is also a rigorous and scholastic step. All the generated profiles have been used to train the model and compare its behavioral changes to enhance the performance even further. After finalizing the model structure and configuration, it can be deployed to an edge device or a cloud service. The role of the edge device can be served by the same electric drive or gateway in a modern electric powertrain (Figure 1). Now that the model has been designed and deployed, it is possible to use it for getting estimations of the case temperature by running inference. At this final point, the model can be further adjusted by running performance tests and considering several evaluation metrics.

Table 2. ML model inputs and target outputs

| Parameter | Symbol |
|---|---|
| **Measured inputs** | |
| Output frequency | $f_d$ |
| Output current | $I_d$ |
| Output power | $P_d$ |
| Ambient temperature | $T_{amb}$ |
| | |
| **Target outputs** | |
| Case temperature | $T_C$ |

## III. Operating cycles of the electric drive

Various static and dynamic input profiles have been used to produce the needed training datasets by feeding the frequency (or speed) and torque to the test drive and motor, respectively (Table 1). Usually, repeating cycles are performed in several applications; however, different scenarios have been tested to explore the model's capability and performance limits. All the generated frequency and motor load profiles have been used for training the ML model; however, some profiles have been kept separate to validate its performance even further. One of these profiles is shown in Figure 4, where the system initially starts from thermal equilibrium, followed by a period of nominal output frequency and motor load. Then, the operation becomes more dynamic with sudden changes in both frequency and load. Under these circumstances, the irregularity of the operation makes temperature monitoring even more challenging for an ML algorithm.

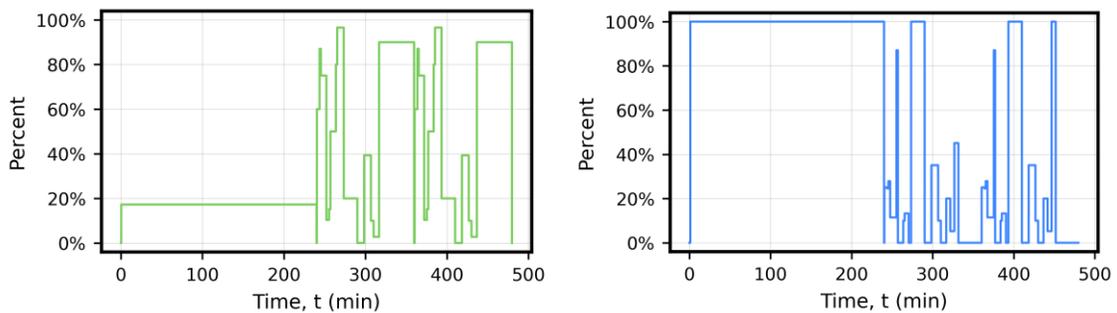

Figure 4. Normalized output frequency (left) and load (right) profiles.

Since the electric drive controls the speed of a motor, its current must remain within its operating limits. As shown in Figure 5, the current stays below its nominal value, although overloading the motor is likely under certain conditions. Another significant input parameter is the ambient temperature, as the model needs to know in which environmental conditions the system is operating. The ambient temperature is typically measured with sensors located in external locations of the electric drive or close to its fan. In the conducted tests, the temperature did not vary drastically; however, there are cases where enclosed systems suffer from inadequate cooling, making temperature inclusion a prerequisite for precise temperature estimation.

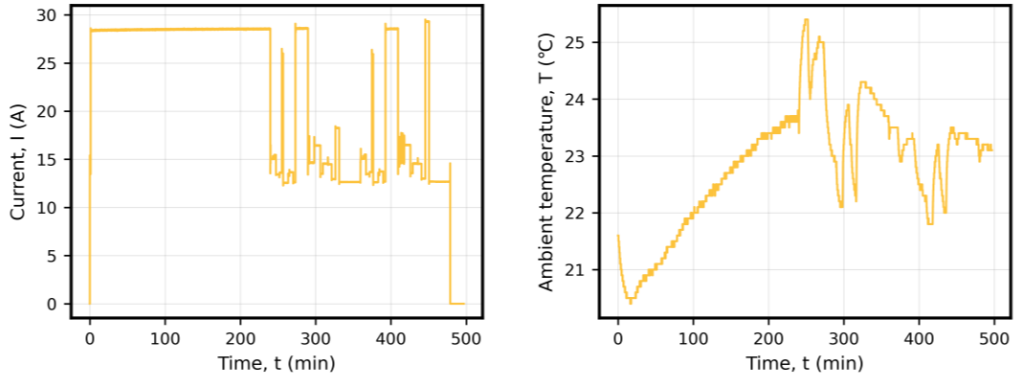

Figure 5. Motor current and ambient temperature rescaled into the range of [0,1].

## IV. Estimating the power module temperature

The trained ML model has been used to estimate the case temperature of the power module after performing the operating cycle in Figure 4. It is worth highlighting that the selected profile was not the best performing; however, it has been selected to emphasize both the model performance and complexity. More specifically, the results of the three developed models have been summarized in Figure 6. All the models perform satisfactorily and maintain an error within the range of +/- 6 oC even under highly dynamic conditions. The error is considerably higher at the beginning of the operation, which is attributed to the initialization of the expanded features of the moving averages. In the second half of the operation, there is a good agreement between all the measured and predicted temperatures.

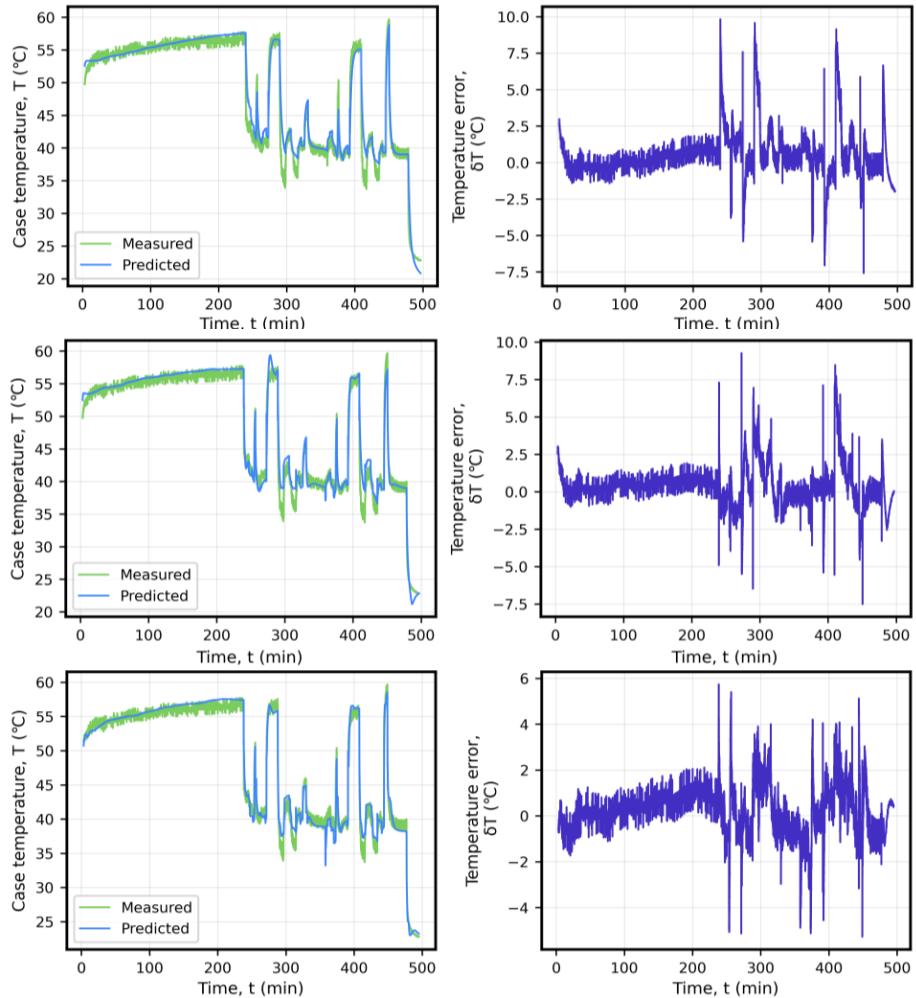

Figure 6. Estimated and measured case temperature with Elastic Net, MLP, and CNNs, from top to bottom.

The comparative analysis of all the methods has shown that linear regression with Elastic Net achieved minimal error. MLP succeeded in keeping the error stretched over the area around zero, whereas CNNs were more biased by exhibiting errors split into two distinct areas. Nevertheless, if the focus is on the second half of the operation after the moving averages have been initialized, then Elastic Net and CNNs were more effective. An error within 10oC is considered exceptional for analytical or even computational models in several applications. Considering that the tested operating cycle has been highly dynamic, it can be safely concluded that all the investigated models have successfully estimated the case temperature.

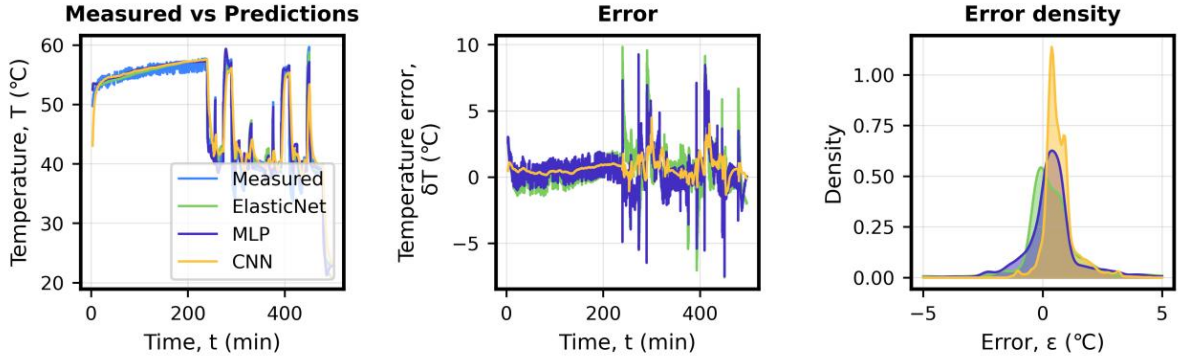

Figure 7. Measurement versus predictions, temperature error, and error density.

In addition to the visual assessment of the performance, several metrics have been used to evaluate the different approaches. Instead of relying on only one profile and concluding that a linear regression may be sufficient, all seventeen profiles have been tested by retraining the ML model and excluding the investigated profile each time. All the results have been summarized in Table 3. It is shown that CNNs exhibited the lowest mean squared error (MSE) and mean absolute error (MAE), whereas the other two models, Elastic Net and MLP, also performed satisfactorily. In terms of complexity, the linear model needed the minimum number of parameters with training and inference times of ~9 times and ~500 times faster than CNNs. Based on the application needs and capacity of the edge device, the most suitable method must be selected, though.

Table 3. Summary of the evaluation metrics under all the tested profiles.

| Model | MSE | MAE | $R^2$ | $l_\infty$ | Training time (s) | Inference time (s) | Parameters |
|---|---|---|---|---|---|---|---|
| ElasticNet | 2.33 | 1.05 | 0.92 | 10.45 | **4.61** | **0.005** | **72** |
| MLP | 2.63 | 1.07 | 0.92 | 8.53 | 49.55 | 0.95 | 8k |
| CNN | 0.94 | 0.68 | 0.96 | 3.74 | 403.84 | 2.46 | 6k |

Furthermore, it is interesting to investigate which of the parameters and their features were the most impactful on the temperature estimation accuracy. A correlation matrix has been employed to select the finally used parameter subset by visualizing hidden patterns; however, at a later stage, the selected parameters have been expanded with more features. As expected, the motor current and its moving average of a 3-minute span were the most impactful (Figure 8). The output power and ambient temperature also played a critical role. All these parameters are directly related to the power losses of the drive and its boundary conditions with the environment; therefore, it is evident that their impact should be high. This is not valid for the expanded features, though, because the impact of the historical drive operation cannot be easily implied heuristically.

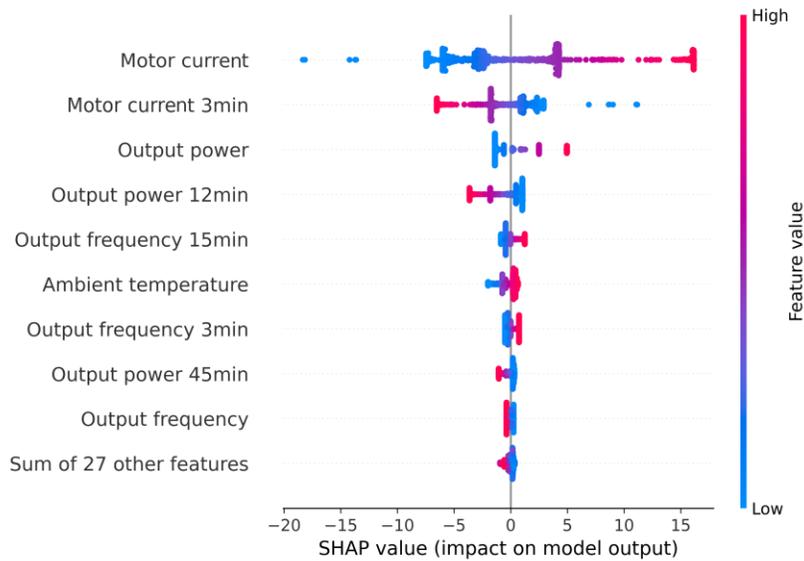

Figure 8. Impact of the parameters on the model output.

## V. Condition monitoring: A case study

Estimating internal drive temperatures with this high accuracy level is extremely valuable, especially if additional sensors are not needed. One of the primary use cases of temperature utilization is condition monitoring. The ML model can provide the anticipated temperature based on past system performance. In contrast, a comparison with measured data from the field can reveal significant findings about its current condition. More specifically, since forced air is used to cool the investigated drive, the air outlet has been partially blocked after 3 hours of operation in a totally new operating cycle (Figure 9). The error between the measured and predicted temperatures becomes large at the moment of the blockage and remains large for the rest of the cycle.

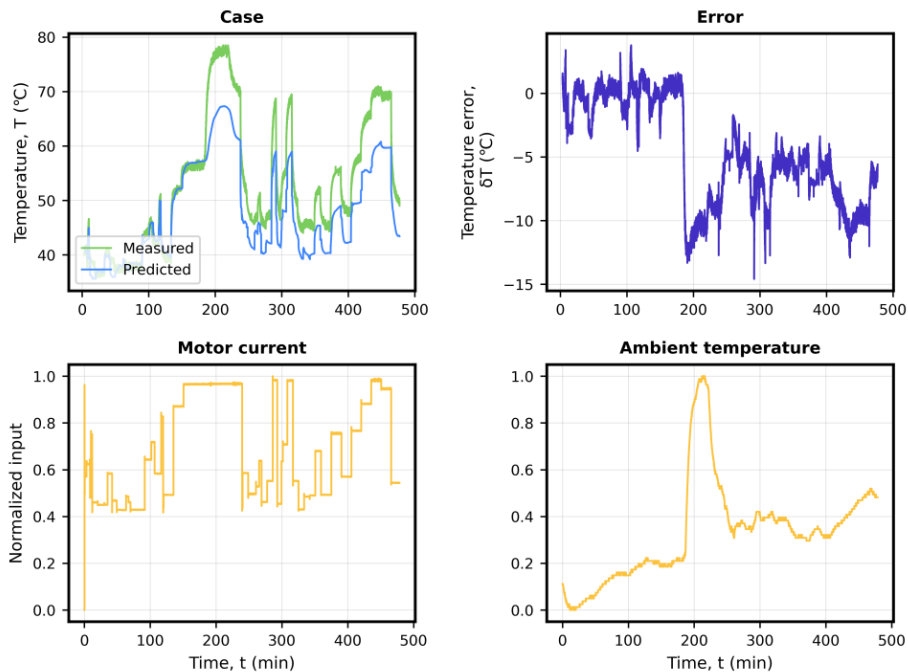

Figure 9. Model performance with partially blocked air outlet after about 3 hours of operation.

This is a realistic use case because industrial systems usually suffer from high temperatures because of dust buildup on the fins or around the vents. The change in the temperature error can signal an alert about the harsh operating conditions, thus preventing a fatal failure from taking place in the first place. A particular remark here is that the injected failure also impacts the predicted temperature. This behavior is attributed to the fact that the ambient temperature used as a model input is measured at a location close to the air outlet, thus being impacted by the fin blockage.

## VI. Summary

Industry digitalization has brought significantly powerful embedded systems that are changing almost every traditional aspect. Training and running ML models on edge is now feasible, bringing promising results. In this regard, this article demonstrated a data-driven thermal model that used data already residing in most modern electric drives and predicted the case temperature of a power module. Conventional linear models (Elastic Net) and deep neural networks (MLP, CNNs) have been investigated for temperature estimation under static and dynamic operating profiles. All models performed satisfactorily but exhibited different characteristics. A compromise based on the application must be made for selecting the simplest algorithms against the most complex but accurate ones. The condition monitoring of electric drives and every other industrial system is about to change drastically due to the high effectiveness of ML in detecting abnormalities. Human intervention may remain high in the coming years. However, as ML becomes more confident by accumulating more data, higher independence in the decision-making process will be the outcome of this exploration.